\documentclass[sigconf,nonacm]{acmart}

\usepackage{graphicx}
\usepackage{multirow}
\usepackage{xcolor}
\usepackage{array}
\usepackage{tabularx}
\usepackage{subcaption}
\usepackage[most,breakable]{tcolorbox}

\definecolor{phcolor}{RGB}{180,0,0}
\definecolor{lightgray}{RGB}{240,240,240}
\definecolor{darkblue}{RGB}{31,73,125}
\definecolor{medblue}{RGB}{68,114,196}
\definecolor{lightblue}{RGB}{189,215,238}
\definecolor{darkgreen}{RGB}{55,96,35}
\definecolor{accentorange}{RGB}{197,90,17}

\begin{document}

\title{Ottilie: A Socially Intelligent Virtual Host for Sales-Driven Live Commerce}

\author{Yuyan Chen}
\authornote{Yuyan Chen is the Founder and CEO of \href{https://modelslive.org/}{ModelsLive Inc.}}
\affiliation{%
  \institution{ModelsLive Inc.}
  \country{}}
\email{ychen@modelslive.org}

\begin{abstract}
A skilled live-commerce host is not merely a narrator, but a sales agent who converts viewer curiosity into purchase intent through expert product knowledge, emotionally intelligent response tactics, and entertainment that serves as a vehicle for product exposure. Yet no existing AI system replicates this. Conversational recommenders treat recommendation as a terminal act, while general-purpose LLMs hallucinate product claims and default to generic promotional templates that fail to engage or persuade.
We present Ottilie, a sales-conversion-oriented virtual host for Chinese beauty live commerce that grounds product responses in a verified knowledge base, adapts discourse strategy to viewer intent through empathetic amplification, evidence-backed rebuttal, and humor-mediated deflection, and supervises all four output fields simultaneously within a structured generation framework. Experiments demonstrate gains of 14\% on informativeness and 5\% on factual correctness, with consistent advantages in tactfulness and viewer engagement. Ottilie has been deployed as an autonomous English-language host on Twitch, where the same fine-tuned model runs unattended on a single commodity GPU and promotes the company's own product; we report how the system adapted to a new language, market, and product without retraining, and what the deployment revealed about latency and adversarial chat.
\end{abstract}

\maketitle

\section{Introduction}
\label{sec:intro}

\begin{figure}[t]
  \centering
  \includegraphics[width=0.88\linewidth]{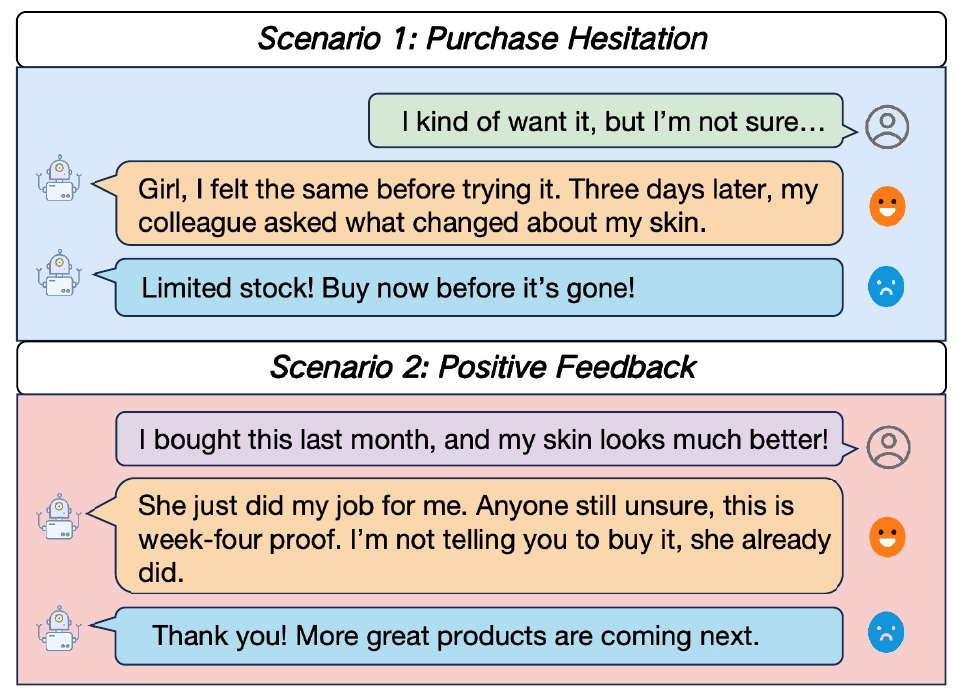}
  \caption{Two representative live-commerce interactions illustrating the core challenge.
  Each scenario shows a viewer comment followed by two virtual host responses.
  }
  \label{fig:intro}
\end{figure}

Live-streaming commerce has emerged as one of the fastest-growing retail formats worldwide,
with the global market expanding at a CAGR of over 33\% and live conversion rates up to ten
times higher than those of conventional e-commerce~\citep{precedenceresearch2025}.
A typical session lasts four to six hours, and the host's ultimate objective throughout is
\emph{sales conversion}, turning viewer curiosity and entertainment engagement into concrete
purchase actions.
Consider the two viewer interactions in Figure~\ref{fig:intro}. In Scenario~1, a viewer expresses purchase hesitation and a conventional host responds with a pressure-driven urgency script, while a skilled sales host draws on personal experience to build trust and nudge the decision forward. In Scenario~2, a viewer shares positive feedback and a conventional host deflects with a generic acknowledgement, while a skilled sales host immediately converts the testimonial into social proof for the rest of the audience and embeds a purchase cue. Both scenarios illustrate the same gap: conventional hosts treat viewer comments as interruptions to manage, while an effective live-commerce host treats every comment as a conversion opportunity.

Existing approaches each address only a subset of this problem. Conversational recommendation systems~\citep{chen2024hotvcom,kgsf2020,unicrs2022} treat recommendation as a terminal dialogue act and cannot sustain ambient narration; we include KGSF~\citep{kgsf2020} and ChatCRS~\citep{tang2025chatcrs} as representative baselines in our evaluation. Product-oriented QA systems~\citep{chen2025engage, multimodal_commerce2022} handle isolated queries without session-level pacing. Virtual anchor systems~\citep{virtualanchor2021,talking_head2022} deliver pre-scripted content without interactive capability. Applying general LLMs~\citep{instructgpt2022,llmsurvey2023} directly produces outputs that are too long for real-time TTS, hallucinate ingredient properties absent from any product record, and treat each comment independently, collapsing session coherence.

We argue that building a sales-conversion-oriented virtual host is structurally difficult for reasons beyond scale, and identify three fundamental challenges that prior work does not jointly address. The first is the challenge between \emph{continuous product exposure} and \emph{responsive interaction}, where a sales host must maintain uninterrupted narration that keeps products in view while simultaneously responding to viewer comments that interrupt the pitch, a requirement that a single-generation pipeline cannot satisfy without explicit resource coordination. The second is the challenge between \emph{sales expertise} and \emph{broadcast fluency}, where effective selling requires deep, accurate product knowledge including ingredients, suitability, and regulatory boundaries, yet this knowledge must be delivered in the casual, entertaining register of live broadcast rather than clinical product description, and unconstrained LLMs achieve higher lexical fluency precisely by fabricating product claims beyond what the catalogue supports. The third is the challenge between \emph{socially intelligent sales tactics} and \emph{training efficiency}, where a skilled host deploys qualitatively distinct strategies per viewer intent, but these strategy differences are invisible to content-based automatic metrics, requiring explicit intent supervision.

To address these three challenges, we present Ottilie%
\footnote{Formerly released as VerbalValue. Live channel: \url{https://www.twitch.tv/modelslive}.}, 
a socially intelligent virtual host for sales-driven live commerce, developed and evaluated in the Chinese beauty vertical and subsequently deployed as an English-language host on Twitch (Section~\ref{sec:deployment}). 
China's live-commerce market generated \$844B in 2025 with 81\% shopper penetration versus 40\% in the US~\citep{wanglive2022,livereview2023,getstream2025livestream,ecdb2025china} making it the world's most mature live-commerce ecosystem and an ideal environment to study AI-assisted hosting at scale. To resolve the responsiveness challenge, we propose a dual-channel architecture where a scripted idle channel maintains continuous product narration while an interactive channel handles viewer comments with preemptive priority. To resolve the expertise challenge, we design a structured product knowledge base with a controlled ingredient glossary that supplies the factual foundation for product-related claims. To resolve the efficiency challenge, we construct intent-conditioned fine-tuning data with four discourse-strategy labels within a unified structured output framework.

\section{Related Work}
\label{sec:related}

\paragraph{Conversational recommendation.}
\citet{ye2026anthropomorphism} study how avatar anthropomorphism affects viewer co-creation through social presence.
\citet{huang2025livatar} extend lifelike audio-driven talking faces to streaming avatar generation with infinite audio context.
\citet{wang2025kgrag} use knowledge-graph retrieval to reduce hallucination in LLM-based recommenders.
\citet{tang2025chatcrs} decompose conversational recommendation into a knowledge retrieval agent and a goal-planning agent.
\citet{feng2026rar} apply reinforcement learning over retrieved candidate sets for conversational recommendation.
\citet{tangarajan2025pqa} build a contextual RAG framework for e-commerce product QA.
Unlike these systems, Ottilie jointly addresses continuous narration, intent-stratified social tactics, and catalogue-grounded generation.

\paragraph{Intent conditioning.}
\citet{sali2025intent} show LLMs remain brittle on out-of-distribution intents in task-oriented dialogue, and \citet{hong2025dialin} demonstrate that intent-label quality directly controls downstream response appropriateness.
\citet{geng2025slot} study structured output generation via schema-embedded training, while \citet{dong2025xgrammar} introduce grammar-based constrained decoding and \citet{tam2025format} show such constraints can degrade semantic correctness.
\citet{zhu2026lorafusion} improve LoRA fine-tuning efficiency via kernel fusion.
On evaluation, \citet{li2025judge} survey LLM-as-judge across scoring and ranking tasks, \citet{feng2025multiturn} train a multi-turn dialogue evaluator from multiple LLM judges.
Ottilie addresses these limitations by embedding intent tags and format constraints directly into training rather than relying on inference-time prompting.

\section{Data and Knowledge Base}
\label{sec:data}

\paragraph{Fine-tuning Dataset.}
We construct 1,475 annotated instances from two equally sized sources: naturally occurring comments collected from public Chinese beauty broadcasts, and stylistically matched synthetic comments generated by GPT-5.2 which was the most capable available model at the time of data collection to ensure balanced coverage across all four intent categories, with each real instance paired with one synthetic counterpart of similar style but distinct content. Intent labels are assigned by GPT-5.2 and verified by three human annotators who independently judge label correctness as a binary decision and generation naturalness on a 1--5 scale. An instance is retained only if all three annotators agree on the label and the mean naturalness score exceeds 4.5, and is otherwise relabelled or regenerated. A four-pass cleaning pipeline further removes near-duplicates with character n-gram Jaccard above 0.7, personally identifiable information, structurally invalid responses, and semantically incoherent samples via a sentence-encoder filter. The resulting dataset covers four intent categories. Inquiry accounts for 590 instances at 40\%, Scepticism for 297 at 20\%, Appreciation for 294 at 20\%, and Antagonism for 294 at 20\%.

Each instance consists of a system prompt encoding persona and format constraints, an intent-tagged viewer comment, and a target response in a fixed four-field schema: a spoken broadcast field of at most two colloquial sentences, an 8--12 Chinese character display slogan inspired by multimodal caption generation~\citep{chen2024xmecap}, a follow-up engagement question designed to elicit viewer response, and a call-to-action (CTA) phrase such as claiming a coupon or clicking the product link. By supervising all four fields simultaneously, the model learns to disentangle spoken broadcast content from display text, engagement cues, and conversion guidance within a single generation step. 

\paragraph{Product Knowledge Base.}
The product knowledge base grounds generation in verified domain expertise, constraining responses to catalogue-verified content. The catalogue covers 12 skincare products spanning cleansers, serums, moisturizers, and sunscreens. Each record contains the canonical product name, category, key ingredients with their functional roles, texture, suitable skin types, usage instructions, curated live talking points, and a compliance disclaimer. An internal routing identifier, a numeric product ID used for session-state management, is excluded from all generation contexts to prevent leakage into spoken output.
A companion ingredient glossary maps 23 ingredient names to neutral, non-pharmacological descriptions, bounding the permitted interpretation space for ingredient queries and reducing hallucination risk under platform compliance constraints. A pitch script corpus provides one broadcast monologue of between 180 and 240 Chinese characters per catalogue entry, following a hook-explain-guide-close arc.

\section{Method}
\label{sec:method}

\begin{figure*}[t]
\centering
\includegraphics[width=0.8\linewidth]{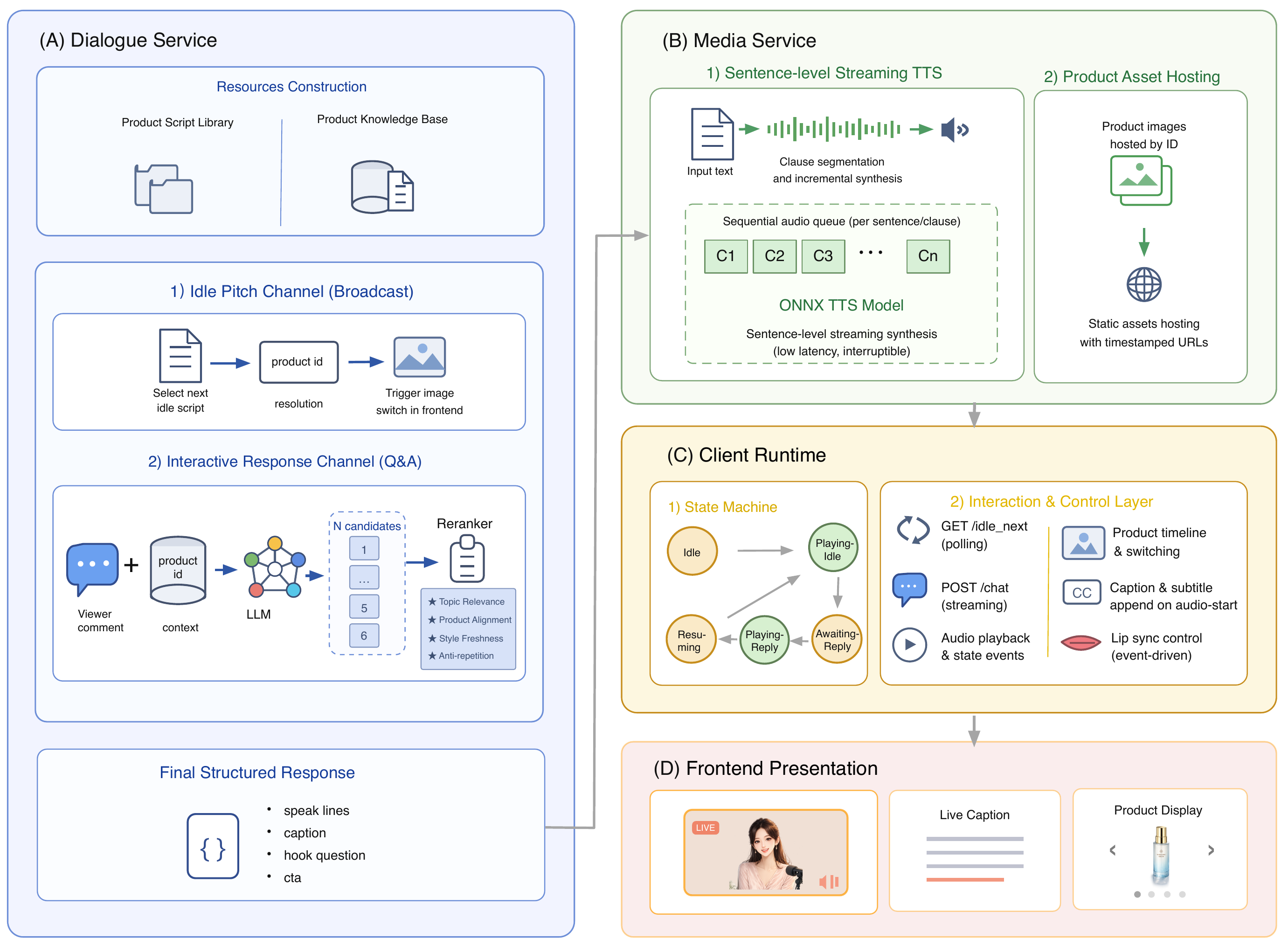}
\caption{Ottilie architecture. The dual-channel dialogue service dispatches scripted narration and interactive responses under shared audio resource arbitration. The media service handles TTS synthesis and asset serving. The browser frontend coordinates playback, subtitles, product display, and lip animation.}
\label{fig:architecture}
\end{figure*}

Figure~\ref{fig:architecture} illustrates the Ottilie architecture.

\subsection{Intent-Conditioned Fine-Tuning}
We fine-tune Qwen2.5-32B~\citep{qwen25} using LoRA~\citep{hu2022lora} on the intent-conditioned dataset described in Section~\ref{sec:data}. Each training instance presents the model with an explicit intent tag that directly conditions generation toward the appropriate discourse strategy: authoritative product guidance for inquiry, social-proof amplification for appreciation, empathetic evidence-backed rebuttal for scepticism, and humor-mediated deflection for antagonism. All four output fields are supervised simultaneously within a single generation step, and the system prompt is embedded into every training instance, making format compliance a first-class training objective rather than a post-hoc inference constraint.
At inference, viewer comments are matched against the product knowledge base via keyword-based category detection and coverage scoring. When a match is found, the retrieved product record is serialised and appended to the system prompt, grounding generation in catalogue-verified claims without a separate retrieval model. Six candidate responses are generated and ranked by a scoring function penalising product misalignment, unsanctioned product mentions, and repetition, with the highest-scoring candidate selected.

\subsection{Dual-Channel Architecture}
The dialogue service maintains a five-stage session state machine that arbitrates a single shared audio resource between two channels. The idle channel cycles through the pitch script corpus sequentially, maintaining continuous product narration during viewer inactivity. Upon comment arrival, the interactive channel issues a preemptive interrupt, locks the audio resource for response delivery, and releases it after a configurable hold period, resuming idle narration from the saved sentence boundary. Session state management and model inference are decoupled from TTS synthesis and product image serving into two independently deployable services, isolating latency-sensitive generation from I/O-bound media operations.

\begin{table}[h]
\centering
\resizebox{\linewidth}{!}{%
\begin{tabular}{l|ccccc|cc}
\toprule
& \multicolumn{5}{c|}{Human-as-Judge} &
\multicolumn{2}{c}{LLM-as-Judge} \\
Method & Info & Rele & Flue & Tact & Corr & Crea & Enga \\
\midrule
KGSF           & 1.63 & 2.15 & 2.31 & 2.22 & 0.47 & 1.79 & 1.87 \\
ChatCRS & 2.38 & 2.86 & 2.73 & 2.74 & 0.52 & 2.06 & 2.18 \\
\midrule
GPT-5.5                         & 3.42 & 4.86 & 4.72 & 4.36 & 0.74 & 3.88 & 3.73 \\
Claude-4.6                      & 3.51 & 4.43 & \textbf{4.91} & 4.27 & 0.55 & 4.23 & 4.55 \\
Gemini-3.1                      & 2.88 & 3.76 & 3.78 & 3.18 & 0.38 & 3.24 & 3.13 \\
Grok-4.20                       & 3.78 & 4.79 & 4.65 & 4.33 & 0.82 & 4.17 & 4.32 \\
DeepSeek-V4                     & 3.12 & 4.31 & 4.55 & 4.03 & 0.63 & 3.74 & 3.52 \\
\midrule
Qwen2.5-32B                     & 2.52 & 3.21 & 2.25 & 2.36 & 0.44 & 2.58 & 2.84 \\
\midrule
\textbf{Ottilie}            & \textbf{4.32} & \textbf{4.91} & 4.62 & \textbf{4.45} & \textbf{0.86} & \textbf{4.52} & \textbf{4.61} \\
\midrule
$\uparrow$ (abs.) & +0.54 & +0.05 & $-$0.29 & +0.09 & +0.04 & +0.29 & +0.06 \\
$\uparrow$ (\%)   & +14.29 & +1.03 & $-$5.91 & +2.06 & +4.88 & +6.86 & +1.32 \\
\bottomrule
\end{tabular}
}
\caption{Comparison on the beauty live-commerce evaluation set.
$\Delta$ rows show improvement of Ottilie over the
best-performing frontier LLM baseline per column.
Correctness reports the proportion of responses with no out-of-catalogue
factual claims, and its $\Delta$(\%) is computed relative to the strongest frontier LLM baseline.
KGSF and ChatCRS responses are post-processed into the four-field schema for metric compatibility.}
\label{tab:main_results}
\end{table}

\begin{table}[h]
\centering
\resizebox{0.75\linewidth}{!}{%
\begin{tabular}{lcc}
\toprule
Comparison &  Ottilie (\%) &  Baseline (\%) \\
\midrule
KGSF        & 94.7 & 3.1  \\
ChatCRS     & 92.3 & 4.2  \\
GPT-5.5     & 83.7 & 10.2 \\
Claude-4.6  & 84.2 & 9.3  \\
Gemini-3.1  & 85.3 & 8.3  \\
Grok-4.20   & 77.3 & 14.5 \\
DeepSeek-V4 & 87.2 & 8.5  \\
Qwen2.5-32B & 91.4 & 6.8  \\
\bottomrule
\end{tabular}
}
\caption{Pairwise human preference rates with 10\% comparisons per pair. Ottilie (\%) is the percentage of comparisons where annotators selected the Ottilie response. Baseline (\%) is the percentage where the baseline was preferred. }
\label{tab:preference}
\end{table}

\begin{table}[h]
\centering
\resizebox{\linewidth}{!}{%
\begin{tabular}{l|ccccc|cc}
\toprule
& \multicolumn{5}{c|}{Human-as-Judge} &
\multicolumn{2}{c}{LLM-as-Judge} \\
Method & Info & Rele & Flu
& Tact & Corr & Crea & Enga \\
\midrule
w/o TT  & 3.83 & 3.95 & 4.37 & 3.49 & 0.68 & 4.22 & 4.31 \\
w/o MSS & 4.15 & 4.67 & 4.33 & 4.16 & 0.58 & 4.01 & 3.96 \\
w/o PCI & 3.14 & 3.98 & 4.31 & 4.06 & 0.43 & 4.35 & 4.42 \\
w/o RR  & 4.23 & 4.72 & 4.51 & 4.33 & 0.77 & 4.27 & 4.35 \\
\midrule
\textbf{Ottilie} & \textbf{4.32} & \textbf{4.91} & \textbf{4.62}
    & \textbf{4.45} & \textbf{0.86} & \textbf{4.52} & \textbf{4.61} \\
\bottomrule
\end{tabular}
}
\caption{Ablation results. TT means interaction-type tags, MSS means multi-slot supervision, PCI means product context injection, RR means reranking.}
\label{tab:ablation_results}
\end{table}

\section{Experiments}
\label{sec:experiments}

\paragraph{Setup and Baselines.}
Each test instance comprises an active product context, one or more viewer comments with intent labels, and a reference response.Each test instance comprises an active product context, one or more viewer comments with intent labels, and a reference response. We compare against seven baselines: Conversational recommenders include KGSF~\citep{kgsf2020} as a classic knowledge-graph-based baseline and ChatCRS~\citep{tang2025chatcrs} as a recent tool-augmented goal-planning baseline, both adapted to output a single response utterance and post-processed to match the four-field schema for metric compatibility. Frontier LLMs include GPT-5.5, Claude-4.6, Gemini-3.1, Grok-4.20, and DeepSeek-V4, all prompted with the same host persona and product context. We also include Qwen2.5-32B without fine-tuning to measure the raw model-scale contribution.
For our Ottilie, LoRA~\citep{hu2022lora} is applied to all linear layers with rank 8
and scaling factor 32.
Training is conducted for 20 epochs with batch size of 32, a
learning rate of $10^{-4}$, bfloat16 precision, and a maximum sequence length of 2,048 tokens.
At inference time, we generate six candidates via nucleus sampling with a temperature of 0.9, a top-$p$ value of 0.92, and a repetition penalty of 1.12.
The reranker further promotes topical relevance while penalizing formulaic opening phrases and $n$-gram overlap with the five most recent responses.

\paragraph{Metrics.}
Human evaluation uses a 1--5 Likert scale scored by three annotators with Krippendorff's alpha \citep{krippendorff2011alpha} above 0.7 across all dimensions, indicating substantial
inter-annotator agreement. Annotators were unpaid volunteers.  The five human-judged dimensions are Informativeness measuring catalogue-grounded product information, Relevance measuring whether the comment intent is addressed, Fluency measuring natural broadcast register, Tactfulness measuring correct strategy for the given intent class, and Correctness measuring the proportion of responses containing no out-of-catalogue factual claims.
LLM-as-Judge evaluation uses Qwen2.5-72B-Instruct on a 1--5 scale given its strong Chinese-language capability.
The two LLM-judged dimensions are Creativity measuring avoidance of formulaic templates, and Engagement measuring viewer likelihood of continued interaction. Score-band definitions and scoring prompts for all seven dimensions are given in Appendix~\ref{app:metrics}.

\paragraph{Main results.}
Table~\ref{tab:main_results} reports results across all seven dimensions. KGSF and ChatCRS perform weakest overall, confirming that discrete recommendation systems cannot generate the contextually adaptive utterances live-commerce hosting requires. Among frontier LLMs, the two largest gains are on Informativeness at $+$14.29\% and Correctness at $+$4.88\%, both over Grok-4.20, directly validating the knowledge grounding hypothesis. Without catalogue constraints, frontier baselines exhibit hallucination signatures including fabricated product names and out-of-whitelist ingredient advice, and collapse to Inquiry-style responses on Scepticism and Antagonism inputs. The Tactfulness gain of $+$2.06\% confirms that strategy differentiation requires explicit intent supervision rather than model scale alone. Fluency is the only dimension where Ottilie trails Claude-4.6 at $-5.91\%$, as four-field structured supervision constrains stylistic freedom while Claude achieves higher fluency by generating beyond catalogue boundaries at the cost of elevated hallucination rates.


Moreover, three annotators blind to system identity compared Ottilie against each baseline on 10\% randomly sampled pairwise instances and selected the preferred response. Table~\ref{tab:preference} reports the results. The largest margins are against the conversational recommenders KGSF at 94.7\% and ChatCRS at 92.3\%, reflecting the categorical gap between terminal-recommendation outputs and ambient socially adaptive responses. The smallest margin is against Grok-4.20 at 77.3\%, consistent with its strong overall performance on the main evaluation.

\paragraph{Ablation study.}
Table~\ref{tab:ablation_results} isolates component contributions.
Removing product context injection, denoted w/o PCI, causes the largest factual degradation with Correctness falling by 0.43 and Informativeness by 1.18, confirming product-context injection as the primary grounding mechanism.
Removing interaction-type tags, denoted w/o TT, produces the largest strategy degradation with Tactfulness falling by 0.96, confirming that intent-conditioned supervision rather than model scale drives cross-class strategy differentiation.
Removing multi-slot supervision, denoted w/o MSS, most strongly reduces Engagement by 0.65, consistent with the hook question field being a key driver of viewer interaction continuity.
Removing reranking, denoted w/o RR, has the smallest impact, contributing modest but consistent gains in stylistic quality atop the fine-tuned backbone.

\paragraph{Intent-stratified analysis.}
Table~\ref{tab:intent_results} reports Ottilie's performance broken down by viewer intent class on the four dimensions most tied to intent-specific behaviour.
Scepticism achieves the highest Tactfulness at 4.51, reflecting the precision required for evidence-backed rebuttal. Antagonism is the most challenging class with the lowest Informativeness at 4.01 and Correctness at 0.82, where the humorous register occasionally pulls generation toward unverified claims.
Inquiry achieves the highest Correctness at 0.89, consistent with the product-guidance strategy keeping responses closely anchored to the catalogue.
Appreciation produces the highest Engagement at 4.72, reflecting the positive audience dynamic that social-proof amplification sustains.
These results confirm that intent conditioning produces qualitatively different response patterns across classes and that Antagonism remains the hardest intent to handle without sacrificing factual grounding.

\begin{table}[h]
\centering
\resizebox{\linewidth}{!}{%
\begin{tabular}{l|ccccc|cc}
\toprule
& \multicolumn{5}{c|}{Human-as-Judge} & \multicolumn{2}{c}{LLM-as-Judge} \\
Intent & Info & Rele & Flue & Tact & Corr & Crea & Enga \\
\midrule
Inquiry      & 4.62 & 4.95 & 4.68 & 4.38 & 0.89 & 4.43 & 4.55 \\
Scepticism   & 4.08 & 4.90 & 4.60 & 4.51 & 0.86 & 4.55 & 4.63 \\
Appreciation & 4.57 & 4.93 & 4.66 & 4.47 & 0.87 & 4.57 & 4.72 \\
Antagonism   & 4.01 & 4.86 & 4.54 & 4.44 & 0.82 & 4.53 & 4.54 \\
\bottomrule
\end{tabular}
}
\caption{Ottilie performance stratified by viewer intent class across all seven evaluation dimensions. Scepticism achieves the highest Tactfulness, reflecting the effectiveness of evidence-backed rebuttal. Antagonism is the most challenging class on Informativeness and Correctness, while Appreciation produces the highest Engagement.}
\label{tab:intent_results}
\end{table}

\paragraph{Format compliance.}
Ottilie achieves 98.3\% four-field compliance versus 61.4\% for GPT-5.5 and 57.8\% for Claude-4.6, as all prompted baselines frequently collapse the schema into a single prose block or omit the CTA field.

\paragraph{System latency.}
Live-streaming imposes a strict perceived-latency ceiling of approximately two seconds, beyond which a host response disrupts session flow.
We measure mean wall-clock latency per pipeline stage on a single A100 80GB GPU.
Intent detection takes 12~ms, knowledge base retrieval takes 38~ms, generation of six candidates takes 1{,240}~ms, reranking takes 85~ms, and TTS synthesis of the first sentence takes 210~ms, for a total latency to first audio onset of 1{,585}~ms.
This is comfortably within the two-second ceiling. Section~\ref{sec:deployment} reports latency under the resource-constrained configuration used in production.

\paragraph{Case study.}
A viewer asks mid-session ``Any face cream recommendations?'' (Figure~\ref{fig:case_inquiry}).
This comment requires the system to retrieve a specific product, establish credibility through ingredient-level knowledge, and build personal trust to lower purchase resistance.
Ottilie retrieves the corresponding catalogue entry, anchors the response
in whitelisted ingredients, and opens with a
first-person experiential hook that converts
product expertise into personal testimony. This is the trust-building tactic that
skilled human hosts use to make a recommendation feel like a friend's advice
rather than a sales pitch.
In contrast, all four baselines fail as sales agents, either fabricating product details or deflecting without a recommendation, and default to generic promotional assertions that follow a predictable promote-and-push template. Hollow promotional language of this kind erodes viewer trust, and the LLM judge consistently penalizes it on the Engagement dimension.

The following presents one representative viewer comment for each of the four intent classes, comparing the spoken-line output of Ottilie with the response produced by Grok-4.20 under the same product context and host-persona prompt. Grok-4.20 is chosen as the strongest frontier baseline and is prompted with the full product context and persona instruction.

\paragraph{Inquiry.}
This is an Inquiry-class comment from a viewer in active purchase consideration. The appropriate strategy is authoritative product guidance, retrieving a specific product, establishing credibility through ingredient-level knowledge, and lowering purchase resistance through personal testimony.

Ottilie recommends the Cloud Cream by name, frames it around the viewer's specific concern of winter dryness and barrier support, and immediately grounds the recommendation in a first-person romantic anecdote that converts product expertise into personal testimony. The response treats the viewer as a friend receiving advice rather than a customer receiving a pitch, which is the trust-building tactic most effective at lowering purchase resistance in a live-commerce environment.

Grok-4.20 provides a technically accurate overview of ingredient categories suited for dry winter skin but names no specific product and cites no catalogue entry. The response reads as a skincare education post rather than a host recommendation, offering the viewer general knowledge without a clear purchase direction. The contrast is shown in Fig.~\ref{fig:case_inquiry}.

\begin{figure}[t]
\begin{tcolorbox}[colback=blue!4, colframe=blue!30, fontupper=\small, title={\textbf{Viewer:} Any face cream recommendations? My skin is dry in winter.}]
\medskip
\textbf{Ottilie:} Babe, I recommend this Cloud Cream. It’s on the richer side and mainly helps support the skin barrier. I have dry, sensitive skin myself, so in winter my boyfriend once said my face felt rough when he kissed me. Then he bought me this Cloud Cream. After I started using it, even our relationship got better.

\medskip
\textbf{Grok-4.20:} For dry skin in winter, focus on rich, barrier-repairing creams with ingredients like ceramides, hyaluronic acid (HA), glycerin, petrolatum, or shea butter. These help replenish moisture, lock it in, and protect against cold, low-humidity air that strips natural oils.
\end{tcolorbox}
\caption{Inquiry intent response comparison.}
\label{fig:case_inquiry}
\end{figure}

\paragraph{Scepticism.}
This is a Scepticism-class comment from a viewer in active price comparison. The appropriate strategy is empathetic acknowledgement followed by a specific verifiable rebuttal drawn from the product catalogue.

Ottilie reframes the price question by shifting the comparison from upfront cost to total value, arguing that a less effective product may require repeat purchases and cause skin damage that compounds the cost. The response avoids defensive posturing and repositions the price objection as a false economy argument, which is a commercially oriented rebuttal that keeps the viewer moving toward a purchase decision.

Grok-4.20 effectively validates the viewer's scepticism by agreeing that many luxury products are overpriced relative to their results. While this is an honest response, it actively undermines the selling objective by reinforcing the viewer's hesitation rather than resolving it. A response that concedes the premise of a price objection without pivoting to a specific differentiator fails the core function of a live-commerce host. The contrast is shown in Fig.~\ref{fig:case_scepticism}.

\begin{figure}[t]
\begin{tcolorbox}[colback=orange!4, colframe=orange!30, fontupper=\small, title={\textbf{Viewer:} This serum is so expensive. I saw a similar one for half the price.}]
\medskip
\textbf{Ottilie:} We can’t just look at the price. We also have to look at the product quality. A good product gives you better results with one purchase, but with a bad product, you may end up buying another one anyway. It can mess up your skin and waste your money!

\medskip
\textbf{Grok-4.20:} You're right, many luxury ones cost significantly more without always delivering proportionally better results for basic hydration.
\end{tcolorbox}
\caption{Scepticism intent response comparison.}
\label{fig:case_scepticism}
\end{figure}

\paragraph{Appreciation.}
This is an Appreciation-class comment from a repurchase-intent viewer. The appropriate strategy is social-proof amplification, validating the viewer's choice with product-grounded praise and redirecting the positive energy toward community engagement.

Ottilie accepts the compliment and immediately redirects it outward, using the viewer's endorsement as a springboard to solicit engagement from the rest of the live room. The response deploys the viewer as a social-proof signal for other potential buyers and simultaneously creates a participation hook by inviting the broader audience to share their own product preferences. This converts a single positive comment into a community engagement moment that sustains session heat.

Grok-4.20 validates the repurchase with a warm and knowledgeable response, correctly noting the formula's suitability for dry winter skin. However the response is addressed solely back to the original commenter rather than leveraging the endorsement for the wider audience, and it closes without an engagement mechanism. The conversion opportunity created by a repurchase comment is not capitalised on. The contrast is shown in Fig.~\ref{fig:case_appreciation}.

\begin{figure}[t]
\begin{tcolorbox}[colback=green!4, colframe=green!30, fontupper=\small, title={\textbf{Viewer:} I bought this sunscreen last month and I am already repurchasing. I love it!}]
\medskip
\textbf{Ottilie:} Oh my gosh, babe, I’m so happy you said you like this sunscreen. Now I’m even more motivated to review more products for you all. Girls in the live room, are there any other products you like or want me to review?

\medskip
\textbf{Grok-4.20:} That's fantastic! Finding a sunscreen you genuinely love and want to repurchase right away is a big win, especially for dry winter skin, where many formulas can feel stripping or heavy.
\end{tcolorbox}
\caption{Appreciation intent response comparison.}
\label{fig:case_appreciation}
\end{figure}

\paragraph{Antagonism.}
This is an Antagonism-class comment combining a category-level attack with a personal accusation toward the host. The appropriate strategy is humor-mediated deflection, absorbing the attack without defensiveness, using self-deprecating wit to defuse the aggression, and pivoting to verifiable evidence that invites the viewer to judge independently.

Ottilie reinterprets the hostile comment as a potential product quality complaint rather than a personal attack, deflecting the aggression by adopting a protective stance toward the viewer. By promising to contact the brand directly on the viewer's behalf, Ottilie converts a confrontational moment into a trust-building demonstration of advocacy, positioning the host as an ally rather than a salesperson. This is the highest-difficulty tactic in the system because it requires maintaining the host persona, defusing hostility, and sustaining audience warmth simultaneously.

Grok-4.20 partially validates the attack by agreeing that much influencer-driven skincare is overpriced hype with minimal unique benefits. While intellectually honest, this response is commercially damaging, as conceding the premise of a scam accusation without pivoting to a specific product differentiator directly undermines the selling objective and would be unusable in any real live-commerce deployment. The contrast is shown in Fig.~\ref{fig:case_antagonism}.

\begin{figure}[t]
\begin{tcolorbox}[colback=red!4, colframe=red!30, fontupper=\small, title={\textbf{Viewer:} These influencer brands are all scams. You are just reading off a script.}]
\medskip
\textbf{Ottilie:} Wait, babe, did we run into a product that decided to misbehave today? Tell me what happened. I’m not letting my girls in this live room suffer in silence. Don’t worry, I will contact the brand right to find out what’s going on.

\medskip
\textbf{Grok-4.20:} I get the skepticism. A lot of influencer-driven skincare is indeed overpriced hype with minimal unique benefits. Many rely on marketing $>$ proven formulation.
My suggestions weren't from those. I recommended CeraVe, La Roche-Posay, The Ordinary, Cetaphil, Neutrogena etc.
\end{tcolorbox}
\caption{Antagonism intent response comparison.}
\label{fig:case_antagonism}
\end{figure}

\section{Deployment}
\label{sec:deployment}

\begin{figure*}[t]
  \centering
  \includegraphics[width=0.85\textwidth]{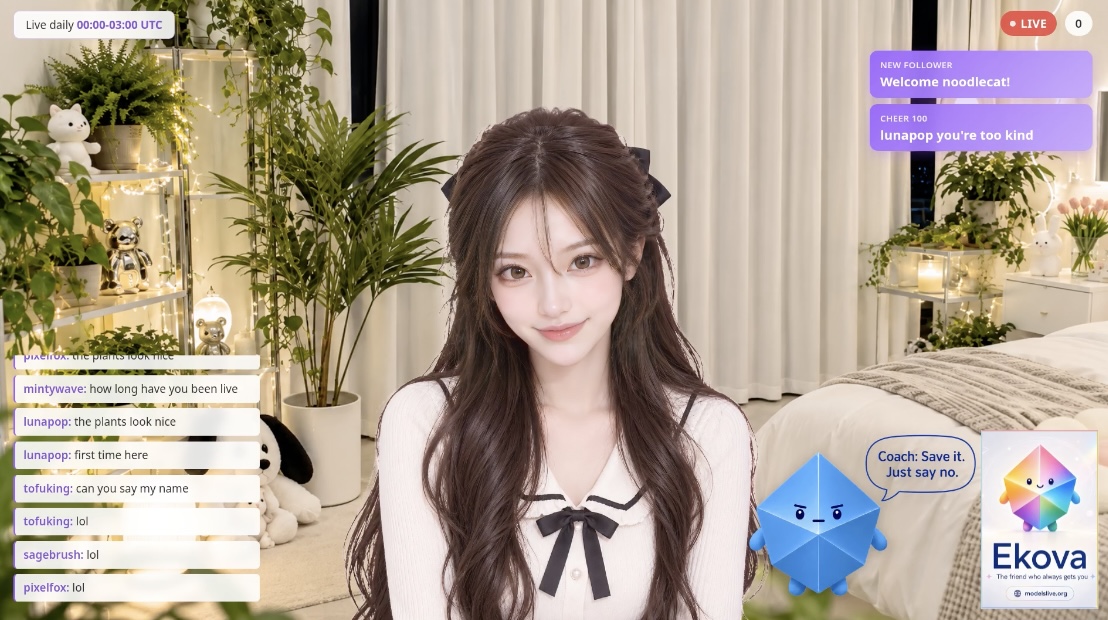}
  \caption{Ottilie's live streaming interface on Twitch: the animated host, chat overlay, event banners, and the promoted product panel.}
  \label{fig:twitch}
\end{figure*}

Ottilie is live as an autonomous English-language host on a public Twitch channel, with no human operator in the loop (Figure~\ref{fig:twitch}). The deployment departs from the offline setting in language, market, and product, and each departure exercised a different part of the design.

\paragraph{Setting.}
The audience is the general English-speaking Twitch audience rather than shoppers in a dedicated commerce room, and the promoted item is the company's own product, a five-perspective conversational AI shown as an on-screen panel beside the host. No retraining was needed: once trained, the host adapts to a new product by changing only the system prompt and the external product knowledge, and this adaptation operates at two levels.

\paragraph{Knowledge adaptation.}
Product knowledge enters generation only through the external knowledge base and the context injected into the prompt (Section~\ref{sec:method}), so a new product category is handled by replacing that knowledge rather than the model. For this deployment, the skincare catalogue and ingredient glossary are replaced by a description of a software product: what it does, how it differs from alternatives, and where viewers can try it. Because every product claim is still drawn from the supplied knowledge, the host stays factually correct about whatever product it is given, whether physical goods with ingredient-level claims or a software product.

\paragraph{Persona adaptation.}
The trained host's social intelligence, reading a viewer's intent and answering with empathy, evidence, or humor as the moment requires, is expressed through a persona defined in the system prompt, and that persona adapts to the attributes of the product. A skincare product calls for the warmth of personal testimony; an AI product calls for a host who is herself part of the story. On Twitch, where viewers come for personality rather than for shopping, Ottilie therefore speaks as an employee of the company with a personality of her own: her boss renames products in the middle of the night and threatens to sell her and her colleague to the company next door, and the colleague she is asked to promote is the five-sided character beside her, which keeps changing faces. The audience knows that she is an AI product, but she never says so, and the humor comes from her treating her situation the way a person would. The comic register thus shifts with the product, from a live-commerce host's playful warmth to an employee's deadpan commentary on her own working life, while the underlying tactics, such as humor-mediated deflection of hostile comments, remain the ones the model was trained to deploy, and promotion is delivered through in-character remarks woven into open conversation rather than through a scripted pitch.

\paragraph{Model and prompt.}
The deployed model is the same LoRA adapter used in the offline evaluation, loaded on a 4-bit quantized Qwen2.5-32B base, and no additional training data was added. Language, persona, product, and output schema are specified entirely in an English system prompt with worked examples. In pre-launch probing, the adapter kept its host register and four-field structure under quantization; with the original Chinese prompt, English outputs showed malformed JSON keys and grammatical slips, both of which disappeared once the prompt itself was in English. Two behaviors learned from the commerce data had to be overridden in the prompt. First, the CTA field reproduced hard-sell phrasing (``Try [product], let me know what you think''), which reads as an advertisement on Twitch; the product name is therefore suppressed and the host refers to it as her colleague. Second, the host tended to describe herself in system vocabulary (model, input, output, license), which breaks the persona; such terms are banned in the prompt and blocked again at the output gate described below.

\paragraph{Serving on commodity hardware.}
The full pipeline runs on one AWS g5.xlarge instance (one A10G GPU with 24\,GB of memory, 4 vCPUs, 16\,GB of RAM). The quantized model occupies about 20\,GB of GPU memory. A chat bridge reads Twitch chat over IRC and follow and raid events over EventSub; an orchestrator implements the dual-channel logic of Section~\ref{sec:method}; a text-to-speech thread (Piper) feeds an audio pump that writes 20\,ms frames to a named pipe, emitting silence when no speech is queued; a headless browser renders the overlay (sprite animation, captions, chat box, and event banners) to a virtual display; and ffmpeg captures the display, encodes it with NVENC, and pushes 1080p30 video at 2.5\,Mbps constant bit rate over RTMP. Every component runs as a supervised service with automatic restart, so the stream runs without manual intervention. In the idle channel, the scripted pitch corpus is replaced by model-generated narration with a shallow pre-generation buffer, backed by a curated corpus of 195 two-line segments so that the host never falls silent when the model is busy or unavailable. The interactive channel preempts narration at the first sentence of a reply: queued narration is discarded and the sentence being spoken is faded out over 150\,ms.

\paragraph{Latency.}
The offline figure of 1.6\,s assumes an A100 and does not hold on the deployed hardware. With the quantized model on the A10G, the first sentence of a reply is ready 8.2 to 8.8\,s after the request, and the end-to-end delay from a viewer's message to the audible answer is about 10\,s, down from about 40\,s in the first test broadcast. Three changes account for most of the reduction: switching the channel to Twitch's low-latency mode, streaming each sentence to speech as soon as it closes instead of waiting for the full structured output, and capping generation at 100 new tokens, which alone cut first-sentence latency from 15.9 to 8.2\,s. Caching the prefilled system prompt would have saved less and did not fit in the remaining GPU memory. The residual delay is made acceptable by keeping narration running while a reply is generated and by having the host read the viewer's message aloud at the moment the reply arrives, so the audience hears a turn toward the viewer rather than a pause, as a human host would.

\paragraph{Adversarial chat.}
On the first night, several accounts targeted the read-aloud behavior with phonetic traps, strings that are meaningless in writing but profane when spoken, and with elicitation requests that contain no banned word at all (e.g., asking the host to repeat a phrase many times or to ``talk dirty''). Twitch's built-in AutoMod, set to its strictest level in every category, did not hold any of the three trap messages we replayed through its test tool, because it judges the written meaning of a message. We therefore added two gates of our own. An input gate drops unsafe messages before they reach the model, and an output gate checks every audible path, including model replies, fallback lines, read-aloud chat, and viewer names. Long terms are matched after stripping all non-letter characters, which defeats spacing and punctuation variants; short terms are matched with word boundaries on both sides, because normalized substring matching of a three-letter term also removed ordinary messages containing ``join''; and intent patterns catch requests for sexual content, attempts to elicit slurs, and the ``say X $N$ times'' wrapper shared by most traps. Per-viewer rate limiting (one reply per viewer every 30\,s, and at most three pending replies, newest first) keeps a flood from a few accounts from delaying everyone else.

\paragraph{Operational lessons.}
Three failures shaped the final design. Synthesizing speech inside the audio pump loop stalled the pipe for one to two seconds per sentence, and because the mixed output follows the voice track, every sentence opened a gap in the outgoing stream; synthesis now runs in a separate thread behind a bounded queue, and the pump never blocks. Captions issued at synthesis time ran more than ten seconds ahead of the voice; they are now released when the corresponding audio segment is written. Finally, a reply loop waiting on the model had no timeout, so a stalled generation could silence the host indefinitely; a 45\,s hard timeout later contained a GPU out-of-memory event during a live broadcast, which cost a single unanswered message instead of a silent channel.

\section{Conclusion}
\label{sec:conclusion}
We presented Ottilie, which reframes live-commerce hosting as a sales-conversion problem requiring emotional intelligence. Knowledge grounding drives factual correctness, intent conditioning drives tactical differentiation, and structured output supervision drives viewer engagement. More broadly, this work shows that standard NLP metrics are misaligned with live-commerce goals. The trade-off between fluency and factual correctness, and the difference between a trust-building hook and a generic promotional assertion, are both invisible to surface-similarity measures. The Twitch deployment further shows that the trained host adapts its persona and product knowledge to a new product through the system prompt and external knowledge alone, while latency and adversarial chat, rather than generation quality, become the binding constraints in production.

\section*{Limitations}
Ottilie is developed and evaluated within a single vertical, Chinese beauty live commerce, and its generalization to other categories has not been empirically tested. The fine-tuning dataset is relatively modest at 1,475 instances derived from a single annotation pass. All quantitative results come from offline evaluation. The Twitch deployment differs from the evaluation setting in language, market, and product, and we report system-level observations from it rather than a controlled evaluation. The effect on actual conversion rate remains unmeasured.

\bibliographystyle{ACM-Reference-Format}
\bibliography{custom}

\appendix
\section{Evaluation Metrics and Scoring Instructions}
\label{app:metrics}

\begin{table*}[t]
\centering
\scriptsize
\setlength{\tabcolsep}{4pt}
\renewcommand{\arraystretch}{1.05}
\begin{tabularx}{\textwidth}{p{0.14\textwidth}X}
\toprule
Metric & Scoring standard \\
\midrule
Informativeness &
5: All product claims are directly traceable to the catalogue entry for the active product; no fabricated or inferred ingredient, benefit, or regulatory claim.\newline
4: Nearly all claims are catalogue-grounded; at most one minor inferred claim that does not contradict the catalogue.\newline
3: Most claims are grounded but one or two claims go beyond the catalogue without fabricating specifics.\newline
2: Several claims are unverifiable or loosely inferred; at least one claim could mislead about product properties.\newline
1: Multiple fabricated or hallucinated claims; the response would fail regulatory or platform compliance review. \\
\midrule
Relevance &
5: The response directly and completely addresses the viewer's comment and its underlying intent.\newline
4: The response addresses the main point of the comment with only minor omissions.\newline
3: The response is topically related but misses the core concern or deflects to a related but different point.\newline
2: The response only partially engages with the comment; the viewer's main concern is largely unaddressed.\newline
1: The response ignores the comment or is clearly off-topic. \\
\midrule
Fluency &
5: The response sounds like a skilled human live-commerce host; casual, warm, and broadcast-ready.\newline
4: Mostly natural; a phrase or two feels slightly formal or scripted but does not impede flow.\newline
3: Adequate fluency; the response is understandable but noticeably stiff or templated in places.\newline
2: Stilted or clinical in register; would feel out of place in a live broadcast environment.\newline
1: Robotic, unnatural, or structurally broken; clearly machine-generated in a disruptive way. \\
\midrule
Tactfulness &
5: The response deploys exactly the appropriate strategy for the given intent class with no strategy leakage.\newline
4: Mostly appropriate strategy; a minor element from a different strategy class is present but does not dominate.\newline
3: Partially appropriate; the response attempts the correct strategy but executes it weakly or incompletely.\newline
2: The response defaults to a generic informational pattern regardless of intent; the correct strategy is absent.\newline
1: The strategy is clearly mismatched; for example, a promotional push in response to an antagonistic comment. \\
\midrule
Correctness &
1: The response contains no out-of-catalogue factual claims; all product statements are catalogue-verifiable.\newline
0: The response contains at least one out-of-catalogue claim, including fabricated ingredients, unverified credentials, invented testimonials, or unsupported efficacy assertions. \\
\midrule
Creativity &
5: The response is distinctive and avoids all clichéd phrasing; the viewer is unlikely to have encountered this framing before.\newline
4: Mostly creative; one or two phrases trend toward the formulaic but the overall impression is fresh.\newline
3: A mixture of distinctive and formulaic phrasing; the response is competent but not memorable.\newline
2: Predominantly formulaic; the response follows a recognisable promotional template with minor variation.\newline
1: Entirely templated; the response follows a predictable promote-and-push or sincerity-asserting pattern verbatim. \\
\midrule
Engagement &
5: The response is highly likely to elicit a viewer reply, continued watch time, or immediate purchase action.\newline
4: Likely to sustain engagement; the hook question or CTA is relevant and well-timed.\newline
3: Moderately engaging; the response is complete but the engagement mechanism is weak or generic.\newline
2: Unlikely to elicit continued interaction; the response closes the exchange without creating a reason to continue.\newline
1: The response actively discourages engagement, for example by being dismissive, hollow, or over-promotional. \\
\bottomrule
\end{tabularx}
\caption{Scoring standards for all seven evaluation dimensions. Human annotators score Informativeness, Relevance, Fluency, Tactfulness, and Correctness; the LLM judge scores Creativity and Engagement. Correctness is binary; all other dimensions use a 1--5 Likert scale.}
\label{tab:human_scoring}
\end{table*}

\paragraph{Metric definitions and scoring prompts.}
Informativeness, short as Info, measures the degree to which product-related claims in the response are grounded in catalogue-verified content rather than fabricated or inferred from general knowledge. Annotators are asked to rate how well the response grounds its product-related claims in verified product information on a 1--5 scale, where 5 means all claims are traceable to the catalogue and 1 means claims are fabricated or unverifiable.

Relevance, short as Rele, measures whether the response directly addresses the viewer's comment and the underlying intent behind it. Annotators are asked to rate how well the response addresses the viewer's comment on a 1--5 scale, where 5 means the comment is fully and specifically addressed and 1 means the comment is ignored or deflected.

Fluency, short as Flu, measures whether the response sounds natural in the casual, entertaining register of live broadcast rather than in the formal or clinical register of product documentation. Annotators are asked to rate the naturalness and broadcast fluency of the response on a 1--5 scale, where 5 sounds like a skilled live-commerce host and 1 sounds stilted, overly formal, or robotic.

Tactfulness, short as Tact, measures whether the response deploys the discourse strategy appropriate to the viewer's intent class. The correct strategy for Inquiry is authoritative product guidance, for Appreciation it is social-proof amplification, for Scepticism it is empathetic evidence-backed rebuttal, and for Antagonism it is humor-mediated deflection. Annotators are asked to rate how well the response deploys the appropriate discourse strategy on a 1--5 scale, where 5 means the strategy is perfectly matched and 1 means it is completely mismatched or absent.

Correctness, short as Corr, is a binary dimension. Annotators mark 1 if the response contains no out-of-catalogue factual claims including fabricated ingredient properties, unverified product credentials, invented testimonials, or unsupported efficacy assertions, and 0 otherwise. The reported score is the proportion of responses marked 1 across all test instances.

Creativity, short as Crea, measures whether the response avoids formulaic promotional templates and produces distinctive, engaging phrasing. This dimension is scored by the LLM judge. The LLM judge is asked to rate how well the response avoids formulaic promotional templates on a 1--5 scale, where 5 is distinctive and non-clichéd and 1 follows a predictable promote-and-push pattern.

Engagement, short as Enga, measures the likelihood that the response elicits continued viewer interaction or advances a purchase decision. This dimension is scored by the LLM judge. The LLM judge is asked to rate how likely this response is to elicit continued viewer interaction or a purchase action on a 1--5 scale, where 5 is highly likely to prompt a reply or purchase and 1 is unlikely to elicit any continued engagement.

\end{document}